%% file: main_revised.tex
\documentclass[letterpaper, 10 pt, conference]{ieeeconf}  

\IEEEoverridecommandlockouts                              

\usepackage{amsmath} 
\usepackage{amssymb}  
\usepackage{url}
\usepackage{graphicx}
\usepackage{cite}
\usepackage{multirow}
\usepackage{comment}
\usepackage{bm}
\usepackage{siunitx}
\usepackage{subcaption}
\usepackage{authblk}
\usepackage[ruled,vlined]{algorithm2e}
\makeatletter
\renewcommand{\Indentp}[1]{%
  \advance\leftskip by #1
  \advance\skiptext by -#1
  \advance\skiprule by #1}%
\renewcommand{\Indp}{\algocf@adjustskipindent\Indentp{\algoskipindent}}
\renewcommand{\Indm}{\algocf@adjustskipindent\Indentp{-\algoskipindent}}
\makeatother
\usepackage[font={small}]{caption}
\usepackage{xcolor}
\usepackage{amsmath}

\usepackage{array}
\usepackage{color}
\usepackage{colortbl}

\title{\LARGE \bf
Sociable and Ergonomic Human-Robot Collaboration through Action Recognition and Augmented Hierarchical Quadratic Programming
}

\author{Francesco Tassi$^{*1,2}$, Francesco Iodice$^{*1,2}$, Elena De Momi$^2$, and Arash Ajoudani$^1$ 
\thanks{* : Both authors contributed equally to this work}
\thanks{$^1$HRI$^2$ Lab, Istituto Italiano di Tecnologia, 16163, Genova, Italy.
{\tt\small francesco.tassi, francesco.iodice@iit.it} 
}
\thanks{$^2$Department of Electronics Information and Bioengineering, Politecnico di Milano, Milan, Italy.}
\thanks{This work was supported in part by the ERC-StG Ergo-Lean (Grant Agreement No. 850932), in part by the European Union’s Horizon 2020 research and innovation programme, Grant Agreement No. 871237 (SOPHIA).}
}%

\begin{document}

\maketitle
\thispagestyle{empty}
\pagestyle{empty}

\begin{abstract}
\input{Abstract_revised}
\end{abstract}


\section{Introduction}
\label{sec:introduction}
\input{Introduction_revised}

\section{Overall Framework}
\label{sec:overall_framework}
\input{ControlFramework_revised}

\section{Ergonomics-Aware AHQP}
\label{sec:proposed_method}
\input{ProposedMethod_revised}

\section{Vision Architectures}
\label{sec:proposed_vision_method}
\input{ProposedVisionMethod_revised}

\section{Experiments and Results}
\label{sec:experiments}
\input{Experiments_revised}

\section{Discussion and Conclusions}
\label{sec:conclusion}
\input{Conclusion_revised}

\bibliographystyle{IEEEtran}
\bibliography{biblio}

\end{document}

%% file: Abstract_revised.tex
The recognition of actions performed by humans and the anticipation of their intentions are important enablers to yield sociable and successful collaboration in human-robot teams. Meanwhile, robots should have the capacity to deal with multiple objectives and constraints, arising from the collaborative task or the human. 
In this regard, we propose vision techniques to perform human action recognition and image classification, which are integrated into an Augmented Hierarchical Quadratic Programming (AHQP) scheme to hierarchically optimize the robot's reactive behavior and human ergonomics. The proposed framework allows one to intuitively command the robot in space while a task is being executed. 
The experiments confirm increased human ergonomics and usability, which are fundamental parameters for reducing musculoskeletal diseases and increasing trust in automation.

\vspace{-.0cm}

%% file: Introduction_revised.tex
\begin{figure*}[!t]
         \centering
         \includegraphics[trim=0cm 0cm 0cm 1.2cm, width=1\textwidth]{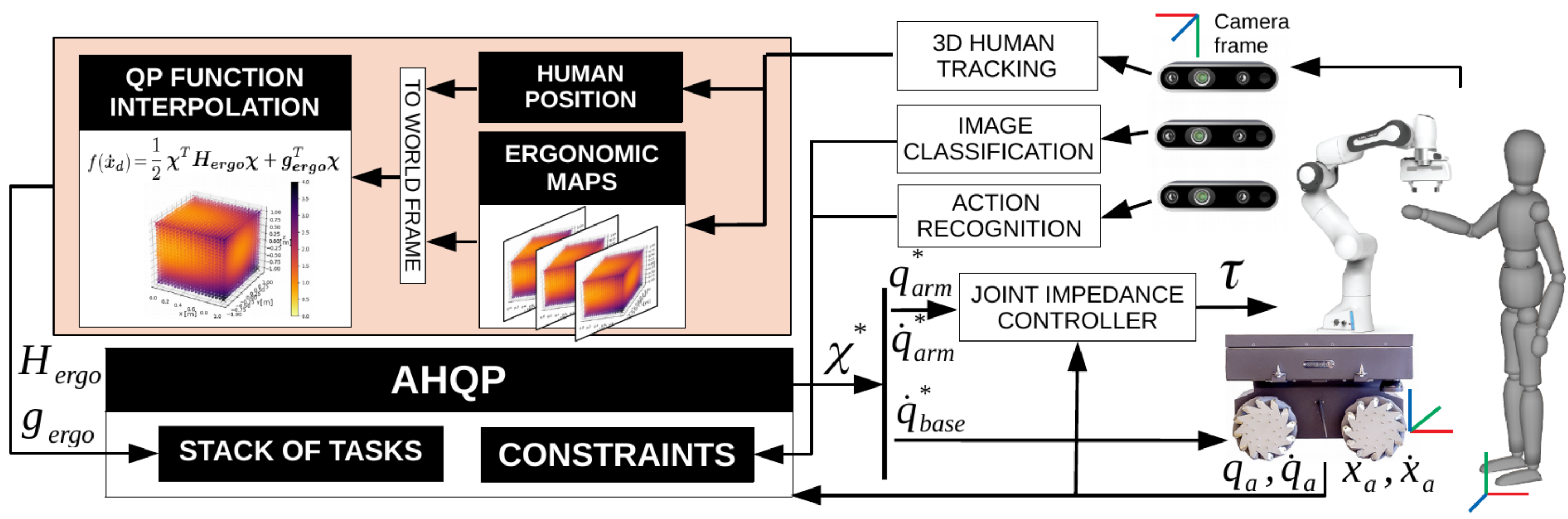}
         \caption{
            Block scheme of the AHQP-based control framework proposed for HRC. The vision system allows to inform the controller about the current behaviour of the human, and to plan the successive tasks based on human intentions (linked to actions recognition).
            This in turn permits ergonomics optimization and successful task completion, thanks to the identification of the optimal solution $\bm{\chi}^*$. 
            }
         \label{fig:blockscheme}
         \vspace{-0.5cm}
\end{figure*}
The number of collaborative robots (Cobots) in human-populated environments has increased exponentially due to their human-comforting properties.
Thanks to strength, rigidity, and precision, Cobots have led to achieve significant developments in Human-Robot Collaboration (HRC), allowing for greater profitability and redistribution of work, thus reducing Musculoskeletal Disorders (MSDs) \cite{ajoudani2018progress}.

This collaboration, however, poses multiple open challenges. For example, it is essential for a Cobot to perceive the intentions of the human and distinguish between a collaborative or non-collaborative behaviour. When collaborating, the robot has to recognize human actions and understand intentions since the operator might interact through the use of multiple tools and objects that the robot should recognize to facilitate the current task.
In addition, HRC is usually conceived within a shared workspace that is fixed in the work environment (e.g., on a workbench), whereas it might be of interest to allow the operator to move the shared workspace while continuing with the task (e.g., reach the next workstation while interacting, reducing cycle times). 

Recently, action recognition has become an important topic, not only for computer vision but also for HRC. Most of the works are based on video processing through deep learning techniques \cite{xiong2020transferable, liu2019deep}.
However, production-related scenarios tend to be oversimplified, and potential complexities due to, e.g., dirty background or similarity between objects or movements, remained largely unexplored. 
Against these limitations, we use the SlowFast \cite{feichtenhofer2019slowfast} network based on a 3D ResNet \cite{tran2018closer} model, which runs on a dataset of action sequences to improve the recognition accuracy.

However, action recognition alone does not ensure higher performances when it is applied to the robot under real circumstances. Indeed, to the authors' knowledge, state-of-the-art methods \cite{xiong2020transferable, liu2019deep, zhang2020recurrent} solve this problem independently from robot control. Our goal is to provide a synergetic framework that fluently integrates action recognition at the control level.
To do so, we require an appropriate control logic capable of fully exploiting Cobots redundancy.
This is possible by defining multiple priority levels of a stack of tasks \cite{Mansard2009_sot}.
Current studies on hierarchical control, range from humanoids \cite{Khatib2008}, and teleoperation \cite{Tassi_Gholami2021} to surgery \cite{Chenguang_Demomi}.
Hierarchical Quadratic Programming (HQP) \cite{Escande} solves multiple Quadratic Programming (QP) problems, establishing strict non-conflicting priorities \cite{Kanoun}.

In HRC, however, it might be beneficial to adjust the desired End Effector's (EE) trajectories online, based on external interactions.
To address this, we proposed in \cite{Tassi2021} an Augmented-HQP (AHQP) scheme for the augmented Inverse Kinematics (IK) problem.
This is based on the creation of a Human-Robot Shared Workspace (HRSW), 
which is the feasible Cartesian region defined by QP constraints, in which HRC can occur. Inside it, thanks to the augmented scheme,
the robot can optimally and autonomously adjust both its reference and actual trajectories (based on constraints and objectives). Thanks to the AHQP, we then introduced in \cite{TassiRCIM} human ergonomics, which we exploit in this work to further integrate human intentions and actions recognition.

We take a large step from our previous works, by considering a moving and continuous interaction point for HRC. Indeed, to the authors' knowledge, all the studies addressing human ergonomics in HRC either employ fixed robots \cite{Busch2018} or consider a precise interaction area (e.g., a workbench for assembly \cite{Heydaryan_HRC_ergo}). Here instead, we provide a HQP-based controller capable of following the human while optimizing ergonomics.
In addition, the proposed method allows dealing with ergonomics efficiently at the control level, by avoiding computationally intensive human models running online, as mostly done in the literature \cite{Wansoo}, and not having to rely on noisy or cumbersome acquisitions of all human joints coordinates (i.e., via camera or motion capture hardware).
Indeed, these methods often require the online solution of the human dynamics/kinematics, which significantly affects computation times. In \cite{Wansoo}, an optimal posture is found in human’s joint coordinates, which then needs to be processed by both the motion planner and by the robot controller. Using the AHQP-based controller instead, only one step is required, reducing cycle times and, as importantly, hierarchically formulating multiple objectives and constraints.

Finally, we integrate a new perception module so that human intentions (e.g., collaborative and non-collaborative) and actions (e.g., walking directions, intended tool use, handover) can be detected and anticipated to create reactive robot responses.
The studied scenario is adaptable to most industrial environments. In particular, first the human delivers a generic object to the Cobot, which is recognized thanks to an image classification module. An action recognition module then informs the Cobot about human intentions, in view of the manufacturing operations to be performed on the same object.
The contributions of this work can be summarily listed as follows.
\begin{itemize}
    \item A model for action recognition on constituent elements of a manufacturing sequence. 
    \item A model for the binary classification on a customized dataset of objects' surfaces.
    \item Human action detection and intention recognition are integrated directly at control level through AHQP.
    \item With respect to \cite{TassiRCIM}, ergonomic HRC is introduced for moving interaction, and an HQP-based HRSW softening strategy is defined.
\end{itemize}
In what follows, we first describe the overall framework in Sec. \ref{sec:overall_framework}. We review in the first part of Sec. \ref{sec:proposed_method} the AHQP framework of \cite{TassiRCIM}, while we present the proposed control method in the second part. In Sec. \ref{sec:proposed_vision_method}, we describe the vision methods implemented. Finally, we validate the proposed structure through the experiments in Sec. \ref{sec:experiments}.

%% file: ControlFramework_revised.tex
The idea of the overall framework is to identify human actions and intentions during HRC, acting on the Cobot to accommodate them. Indeed, while most techniques consider a fixed HRC area, and require the operator to adapt to the robot, our goal is to improve the task's degree of flexibility, allowing the human to move in space (e.g., reach the next workstation, grab another tool or item for assembly), while the robot follows the human and optimizes ergonomics.
In the considered scenario, an object is first handed to the robot, identified by image classification and finally grasped.
Successively, the operator can decide which tool to grab, to work on the object (i.e., drill, polisher), and the robot can recognize the intention via the action recognition module and act accordingly (i.e., properly orienting the object's surface).
Furthermore, the robot uses the detection of additional human motions to interpret intents and follow the human in space, assisting him/her in completing the primary task.
This will be detailed in Sec. \ref{sec:exp4}.
Fig. \ref{fig:blockscheme} depicts the block scheme of the proposed framework, in which we employ three RGB-D cameras. The first is mounted on the mobile base to provide clear visibility of the EE. This allows to identify and correct the object's orientation during the activity. The other two are set at a distance that frames the whole working area, preventing occlusions to the greatest extent. One is used for action detection and the other for 3D human tracking, which is required to update online the position and orientation of the ergonomics map defined in Sec. \ref{sec:Human Ergonomics}.
\vspace{-.1cm}

%% file: ProposedMethod_revised.tex
\vspace{-.1cm}
\subsection{Augmented Kinematics}
\label{sec:Augmented IK}
The classic hierarchical structure for $k \in \{1, \dots p\}$ priority levels, with decreasing task importance down to $p$ writes as:
\begin{align} \label{eq:genericQP_complete}
    \begin{split}
     & \min_{\bm{\chi}} \, \frac{1}{2}\, ||\bm{A_k \chi}-\bm{b_k}||^2 \\
     s.t.\;\;  & \bm{C_1 \chi \leq d_1},  \;\; \hdots{}, \;\;  \bm{C_k \chi \leq d_k} \\
  & \bm{E_1 \chi = f_1}, \;\; \hdots{}, \;\;  \bm{E_k \chi = f_k},
  \end{split}
\end{align}
$\bm{\chi} \scriptstyle \in \displaystyle \mathbb{R}^{s}$ being the generic $s$-dimensional variable to optimize, $n_{t_{k}}$ is the dimension of the $k$-th task, subject to $n_{e_{k}}$ and $n_{i_{k}}$ equality and inequality constraints respectively, through the matrices $\bm{A}_{k} \in \mathbb{R}^{n_{t_{k}}\times s},\bm{C}_{k}\in \mathbb{R}^{n_{i_{k}}\times s},\bm{E}_{k} \in \mathbb{R}^{n_{e_{k}}\times s}$ and vectors $\bm{b}_{k}\in \mathbb{R}^{n_{t_{k}}},\bm{d}_{k}\in \mathbb{R}^{n_{i_{k}}},\bm{f}_{k} \in \mathbb{R}^{n_{e_{k}}}$.
To establish the strict hierarchy between successive tasks, it is necessary that the task at a lower priority level $k-1$ is projected in the nullspace of the task holding a higher priority $k$. This is achieved thanks to the condition \resizebox{.3\hsize}{!}{$\bm{A_{k-1} \chi} = \bm{A_{k-1} \chi^*_{k-1}}$} demonstrated in \cite{Kanoun}. To account for this condition in \eqref{eq:genericQP_complete}, it is necessary to update the constraints matrices and vectors at each step by formulating $\bm{ E_1} = \bm{0}, \bm{ f_1} = \bm{0}$, up to $\bm{ E_k} = \bm{A_{k-1}}, \bm{ f_k} = \bm{A_{k-1}\, \chi^*_{k-1}} $. 

The IK problem of an $n$-DoF redundant robot is
\begin{equation} \label{eq:min_kyn}
    \min_{\bm{\dot{q}}} \| \bm{J\dot{q}-\dot{x}} \|^2
\end{equation}
where $\dot{\textbf{q}} \in \mathbb{R}^n$ is the desired joint velocity, $\dot{\textbf{x}} \in \mathbb{R}^{m}$ the task space velocity and $\textbf{J(q)} \in \mathbb{R}^{m \times n}$ is the task Jacobian matrix.
The AHQP scheme proposed in \cite{Tassi2021} solves problem \eqref{eq:min_kyn} through the augmented state $\bm{\chi}$, with the desired Cartesian velocity $\bm{\dot{x}_d}\in \mathbb{R}^{m}$ as part of the optimization variable:
\begin{equation} \label{eq:state_augment}
    \bm{\chi} = 
    \begin{bmatrix} 
    	\bm{\dot{q}}\\
    	\bm{\dot{x}_d}
    \end{bmatrix}, \bm{\chi} \scriptstyle \in \displaystyle \mathbb{R}^{s=n+m}.
\end{equation}

Using a Closed-loop IK (CLIK) scheme in augmented form, the augmented IK writes as in \cite{TassiRCIM}:
\begin{gather}
    \min\limits_{\bm{\dot{q}}, \bm{\dot{x}_d}} \Big\lvert\Big\lvert
    \begin{bmatrix}
        \bm{J} & -(\bm{I}+\bm{K_p}\Delta t)
    \end{bmatrix}
    \raisebox{-.15cm}{$\begin{bmatrix} 
    	\bm{\dot{q}}\\ \bm{\dot{x}_d}
    \end{bmatrix}$} - \bm{K_p}( \bm{x}_{d_{t-1}} -\bm{x}_a) \Big\lvert\Big\lvert^2 
    \nonumber \\
    = \min_{\bm{\chi}} ||\bm{A_{clik} \chi}-\bm{b_{clik}}||^2 \label{eq:CLIK}
\end{gather}
where $\bm{x}_a, \bm{x}_d \in \mathbb{R}^m$ are the actual and desired Cartesian poses of the EE respectively, $\bm{K_p} \in \mathbb{R}^{m\times m}$ is the positive-definite diagonal gain matrix responsible for convergence of the error, $\Delta t$ is the control period that is set to $1 \si{ms}$, and $\bm{x}_{d_{t-1}}$ is the value of $\bm{x}_d$ at the previous time instant.
Eq. \eqref{eq:CLIK} complies with the structure of \eqref{eq:genericQP_complete} and can be solved hierarchically through $\bm{A_{clik}} \in \mathbb{R}^{m\times s}$ and $\bm{b_{clik}}\in \mathbb{R}^{m}$.

\vspace{-.1cm}

\subsection{Human Ergonomics}
\label{sec:Human Ergonomics}
\vspace{-.1cm}

Based on our previous work \cite{TassiRCIM}, we hereby describe how to integrate human ergonomics into the hierarchical framework thanks to the augmentation process.
In HRC, it is important to optimize human ergonomics through Cobots, not by merely acting on their reference trajectories to be followed. Indeed, by defining priorities between productivity and human features at the control level, an optimal trajectory can be generated.
However, to achieve this in real-time, a lightweight formulation is difficult to achieve when considering the highly redundant human kinematics computation. Therefore, based on the assumptions made in \cite{TassiRCIM}, we identify an ergonomics map in Cartesian space, simplifying the ergonomics problem and allowing for its tractability in the AHQP framework.

To assess human ergonomics and identify the ergonomics map, a Rapid Entire Body Assessment (REBA) score is considered. Given its highly discrete nature, though, the different contributions that compose the overall score (arms, trunk, legs, and neck) are considered separately and ultimately averaged. This provides continuous variations, useful to distinguish between similar postures that would correspond to the same REBA score.
Dealing with HRC, it is obvious to assume $\bm{x_h} = \bm{x_d},$ for which the coordinates of the human hand $\bm{x_h}$ correspond to the desired EE position, meaning both human and robot are ideally collaborating. This leads to the explicit formulation of the ergonomics score with respect to a subset of the optimization variable as $e_s = f(\bm{x}_d)$,
where the score $e_s \in \mathbb{R}$ can be employed in the HQP.

The other assumption is related to the definition of the ergonomics map. In particular, we assume that two persons (with the same kinematic parameters, center of mass position, and orientation in space) reach for a specific point in space with very similar postures of the overall kinematic chain $\bm{q_h}$, over multiple repetitions \cite{TassiRCIM}. Indeed, despite the high degree of redundancy, this was confirmed by observing multiple subjects and recording their postures, noticing that they all tend to assume a configuration that minimizes the overall human joints activation. This, in turn, provides a measure of overall ergonomics associated with each point in the reachable space.
Ultimately, the optimal ergonomics task is written as a function of $\bm{\dot{x}_d}$ and in QP form as:
\begin{align} \label{eq:min_ergo}
    & \min\limits_{\bm{\dot{q}}, \bm{\dot{x}_d}} \ e_s =  \min\limits_{\bm{\dot{q}}, \bm{\dot{x}_d}} f(\bm{\dot{x}_d}) =  \min\limits_{\bm{\chi}} f_{a}(\bm{\chi}) 
    \\ & = \min\limits_{\bm{\chi}} \frac{1}{2}\, \bm{\chi}^T \bm{H_{ergo}} \bm{\chi} + \bm{g_{ergo}}^T \bm{\chi} \label{eq:Hgergo}
\end{align}
with $\bm{H}_{ergo} \in \mathbb{R}^{s\times s}, \bm{g}_{ergo} \in \mathbb{R}^{s}$ containing the parameters obtained from the mapping. Finally, this is used in the AHQP structure \eqref{eq:genericQP_complete}, to regulate the position in which the robot's EE has to collaborate with the human for the task.

\vspace{-.1cm}

\subsection{Softening of the augmented problem}
\label{sec:constraints}

\vspace{-.1cm}

This section describes the constraints regulating the augmented problem and defines a slack variable useful for constraints softening, necessary to achieve the advantages related to the HRSW.
This control scheme provides the possibility of obtaining $\bm{\dot{x}_d}$ as output from the optimization, avoiding the restrictions of a fixed reference trajectory provided by upstream planning.
Thereby, it is possible to simply act on the Cartesian constraints that identify the HRSW $\bm{x}_{d_{min}} \leq \bm{x_d} \leq \bm{x}_{d_{max}}$, and the controller will provide the optimal solution based on the hierarchy.
To achieve this, however, it is necessary to avoid the infeasibility problems that would arise when the HRSW is modified, and the current value of $\bm{\dot{x}_d}$ lies outside. 
Besides, it is beneficial to provide a degree of flexibility with which the optimization variable can remain outside of its boundaries, at the expense of another task ranked higher in the hierarchy (e.g., optimal ergonomics).
Therefore, a slack variable $\bm{s}$ is added on $\bm{x_d}$:
\begin{gather} \label{eq:constr}
    \min\limits_{\bm{\chi}, \bm{s}} \  \frac{1}{2}\, \| \bm{s} \|^2  \\
    s.t. \ \bm{x}_{\bm{d}_{min}} - \bm{s} \leq \bm{x_d} \leq \bm{x}_{\bm{d}_{max}} + \bm{s}, \nonumber 
\end{gather}
softening the HRSW and avoiding infeasibility. Rewriting as:
\begin{gather}
    \begin{bmatrix}
        \bm{0}_{m\times n} & \Delta t \bm{I}_{m\times m} & -\bm{I}_{m\times m}\\
        \bm{0}_{m\times n} & -\Delta t \bm{I}_{m\times m} & -\bm{I}_{m\times m}
    \end{bmatrix}
        \begin{bmatrix} 
    	\bm{\dot{q}}\\ \bm{\dot{x}_d} \\ \bm{s}
        \end{bmatrix}
    \leq \begin{bmatrix} 
        \bm{x}_{\bm{d}_{max}} - \bm{x_{d_{t-1}}} \\
        \bm{x}_{\bm{d}_{min}} - \bm{x_{d_{t-1}}}
        \end{bmatrix}
    \nonumber\\
    = \bm{C}_{s} \begin{bmatrix} \bm{\chi} \\ \bm{s} \end{bmatrix} \leq \bm{d}_{s},
    \label{eq:constr_slack} 
\end{gather}
where $\bm{C}_{s} \in \mathbb{R}^{2m\times s+m}$, $\bm{d}_{s} \in \mathbb{R}^{2m}$ and the state is further augmented by $\bm{s}$.
Ultimately, as we show in the experiments, we will update the HRSW online to make the robot follow a continuous trajectory dictated by the human.

All the other constraints on $ \bm{\dot{x}_d},  \bm{\ddot{x}_d}$ and $\bm{q}, \dot{\bm{q}}, \ddot{\bm{q}}$, are solved through the inequality parameters $\bm{C_k}, \bm{d_k}$ in \eqref{eq:genericQP_complete} as:
\begin{equation}
    \bm{\chi}_{min} \leq \bm{\chi} \leq \bm{\chi}_{max}.
\end{equation}

\subsection{Person-specific Cartesian mapping}
\label{sec:Human-specific Cartesian mapping}
\vspace{-.1cm}
Considering a realistic HRC task, it is often necessary for the robot to interact with multiple people having various characteristics of height, posture, and overall kinematics. 
To address this, we extend the ergonomics behavior of Sec. \ref{sec:Human Ergonomics} by defining multiple maps of the score tailored to subjects with different kinematics (i.e., one for tall, average, and short subjects, respectively). This further optimizes the ergonomics of the final robot's trajectory, which can now provide a response tailored to each operator.

\subsection{Stack of tasks}
\label{sec:sot}
\vspace{-.1cm}
The overall stack of tasks is composed of three strict hierarchical levels. Highest priority is given to the formulation of the CLIK defined in \eqref{eq:CLIK}, while at the second level we minimize the slack variable $\bm{s}$ responsible for softening the Cartesian constraint on $\bm{x_d}$ \eqref{eq:constr}. Lastly, we prioritize the ergonomics function \eqref{eq:Hgergo} that is responsible for dictating the optimal $\bm{x_d}^*$. This order, aims at the highest satisfaction of human ergonomics while remaining inside the HRSW.

The optimal solution $\bm{\chi}^*$ found, is composed of:
\begin{equation} \label{eq:state_augment_base}
    \bm{\chi}^* = 
    \begin{bmatrix} 
    	\bm{\dot{q}}^*\\
    	\bm{\dot{x}_d}^*
    \end{bmatrix} =
    \begin{bmatrix}
        \begin{bmatrix} 
        	\bm{\dot{q}_{base}}^*\\
        	\bm{\dot{q}_{arm}}^*
	    \end{bmatrix}  	\\
    	\bm{\dot{x}_d}^*
    \end{bmatrix} =
    \begin{bmatrix}
        \begin{bmatrix} 
        	\bm{\dot{x}_{base}}^*\\
        	\bm{\dot{q}_{arm}}^*
	    \end{bmatrix}  	\\
    	\bm{\dot{x}_d}^*
    \end{bmatrix}
\end{equation}
where $\bm{\dot{x}_{base}}^* = ( \dot{x}, \dot{y}, \dot{\theta} )$ are the linear and angular velocities of the mobile base.
While $\bm{\dot{x}_{base}}^*$ is sent to the velocity controlled base, for the manipulator a joint impedance controller is necessary to generate the desired control torques:
\begin{equation} \label{eq:joint_imp_ctrl}
    \bm{\tau} = \bm{K_{q_{d}}} (\bm{\dot{q}_{arm}^* - \dot{q}_a}) + \bm{K_{q_{p}}} (\bm{q_{arm}^* - q_a}) + \bm{g}(\bm{q_a})
\end{equation}
where $\bm{\dot{q}_a} , \bm{q_a} \in \mathbb{R}^n$ are the actual joint velocities and positions respectively, $\bm{q_{arm}^*}$ is obtained from numerical integration of $\bm{\dot{q}_{arm}^*}$, whereas $\bm{K_{q_{p}}}, \bm{K_{q_{d}}} \in \mathbb{R}^{n\times n}$ are the positive definite joint stiffness and damping matrices respectively, and $\bm{g}(\bm{q_a}) \in \mathbb{R}^n$ is the gravity compensation term.

%% file: ProposedVisionMethod_revised.tex
\subsection{Surface Classification Model}
The surface classification model classifies the sides of the object shown in Fig. \ref{fig:Classification_Model}, used for the experiments in Sec. \ref{sec:experiments}. Below, we provide prerequisites and propose a method to use transfer learning, SVM, and ensemble classification.
\subsubsection{Two-Class Linear Support Vector Machine}
The linear support vector machine (SVM) is a supervised machine learning algorithm initially formulated as a binary discriminative classifier. It is trained using a set of labeled instances, called training data, defined as: $\textbf{S}=((x_1,y_1),\cdots ,(x_N,y_N)) \subseteq (\textbf{X}\times \textbf{Y})^{N}$
where $N$ is the number of instances, $\textbf{X} \subseteq \mathbb{R}^{N} $ is the input space and 
$\textbf{Y}=\left\{ 1,-1  \right\} \subseteq \mathbb{R}$ is the output domain. Moreover, we refer to $\textbf{x}_i$ as instances and $\textbf{y}_i$ as their labels.
The objective of SVM is to find the optimal hyperplane:
\vspace{-.3cm}
\begin{equation}
f\left( \textbf{x}\right)= \textbf{w}^{T} \cdot  \textbf{x} + b = \sum_{\mathrm{j=1}}^{N}w_j x_j+b .
\vspace{-.1cm}
\end{equation}
SVM learns the parameters $\textbf{w}$ by solving the problem:  
\vspace{-.2cm}
\begin{align} 
    \begin{split}
&\min\limits_{\textbf{w}, \xi_i} \;\;\ \frac{1}{2} \textbf{w}^{T}\textbf{w} + \mathrm{\sum_{i=1}^{N}}{\xi_i} \\
     s.t.\;\;\;\;  & \xi_i \ge  0  , \hspace{3cm} i= 1, \cdots ,N \\
& \textbf{y}_i \left(  \textbf{w}^{T} \cdot \textbf{x} + b\right)\ge 1-\xi_{i}, \;\;\;\;i= 1, \cdots ,N 
\end{split}
\end{align}
where $\xi_i$ is a slack variable measuring the distance between the margin and the instances $x_i$ lying on the wrong side of the margin, C is the trade-off parameter added to maximize the margin and minimize the classification error.
The SVM solves the unconstrained optimization problem as follows:
\begin{equation}\label{eq:L1_SVM}
    \min\limits_{\textbf{w}} \;\;\ \frac{1}{2} \textbf{w}^{T}\textbf{w} + C \mathrm{\sum_{i=1}^{N}} \max (1 - \textbf{y}_{i}(\textbf{w}^{T}\cdot \textbf{x}_i + b), 0)
\end{equation}
which is known as the L1-SVM primal problem with standard hinged loss. Its differentiable counterpart
\begin{equation}\label{eq:L2_SVM}
    \min\limits_{\textbf{w}} \;\;\ \frac{1}{2} \textbf{w}^{T}\textbf{w} + C \mathrm{\sum_{i=1}^{N}} \max (1 - \textbf{y}_{i}(\textbf{w}^{T}\cdot \textbf{x}_i + b), 0)^{2}
\end{equation}
is known as L2-SVM. It minimizes a higher (quadratic) loss for points that violate the margin.

\subsubsection{Resnet} \label{subsubsection:Resnet}
ResNet \cite{he2016deep} is one of the most powerful deep neural networks,
whose versions (e.g., Resnet-18, Resnet-34, Resnet-50, etc.) only differ in the number of layers.
The central idea of ResNet is the introduction of a so-called "skip connection". that skips two or more layers.
This method causes the residual mapping to allow the stacked layers to fit more easily than the desired background mapping. Also, the stacking of layers does not degrade the network's performance since the original information is not altered and flows directly from layer to layer throughout the network, preventing gradients from disappearing or exploding, improving learning and accuracy.

\subsubsection{Dataset} \label{subsub:Dataset}
The dataset includes 2000 images divided into two classes representing the surface of an object (Smooth and Drilled) uniformly distributed and manually annotated. We collected these images using an RGB camera,
looking for depth variation and background complexity.

\begin{figure}[t]
         \centering
         \includegraphics[trim=2cm .5cm 2cm 1.5cm, width=.9\linewidth]{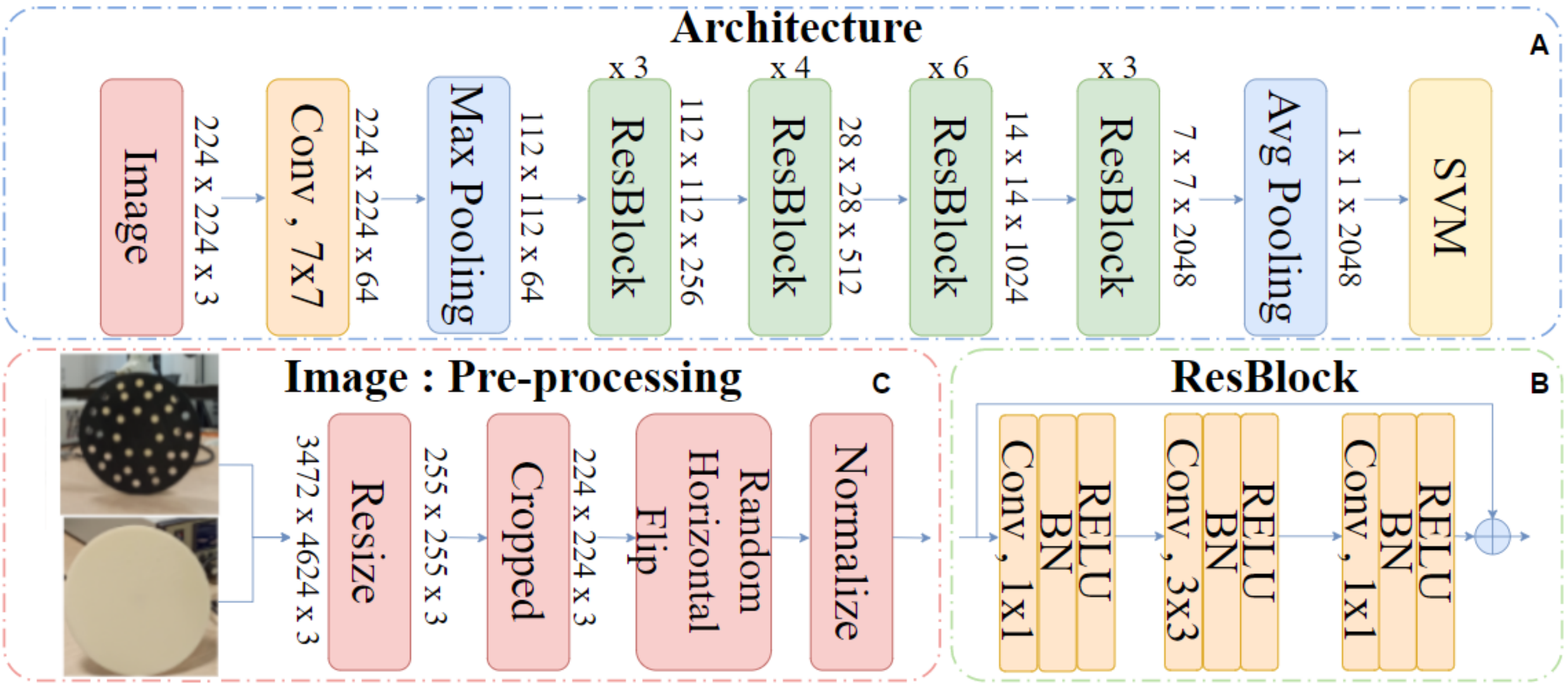}

         \caption{A) Proposed architecture for surface classification based on fusion of ResNet-50 and SVM binary classifier via transfer learning via extractor features. B) Residual blocks with skip connection included in the network. C) Image pre-processing before training.
            }
         \label{fig:Classification_Model}
         \vspace{-.5cm}
\end{figure}
\subsubsection{Description Model} \label{subsub:description_model}
Deep learning models usually need significant amounts of data and tuned architectures to predict with high accuracy. Being our data limited to 2000 samples, we investigate transfer learning techniques, where knowledge is transferred from a different domain.
In particular, we use the Resnet network and the SVM binary classifier. The former, for its simple structure described in \ref{subsubsection:Resnet}, provides better performances than other nets; the latter instead ensures high accuracy in handling small datasets. We choose the Resnet-50 architecture as the backbone for its high accuracy, computational speed, and ease of use, and we build a convolutional network (CNN), a combined model of ResNet-50 and SVM for ensemble classification, implementing transfer learning via extractor features. 
We use the pre-trained ResNet-50 weights on ImageNet as a starting point and freeze them to avoid their modification. This approach avoids back-propagation on the layers and significantly increases velocity during the training phases. Next, we remove the last fully connected layer having as output the number of previously towed classes on ImageNet, and use the features extracted from the average pooling layer to feed a binary SVM classifier (Fig. \ref{fig:Classification_Model}). 
\vspace{-.1cm}

\begin{figure}[!t]
         \centering
         \includegraphics[trim=0cm 0.5cm 0cm 1.5cm, width=0.35\textwidth]{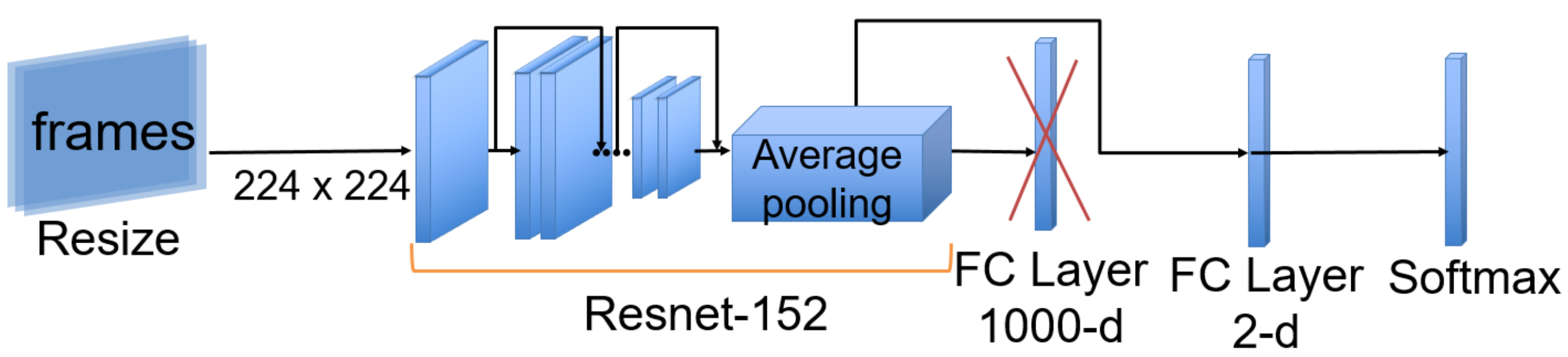}

         \caption{Surface classification model with Resnet-152
            }
         \label{fig:resnet152}
         \vspace{-.2cm}
\end{figure}
\begin{figure}[!t]
         \centering
         \includegraphics[trim=0cm 0.5cm 0cm 1.5cm, width=0.35\textwidth]{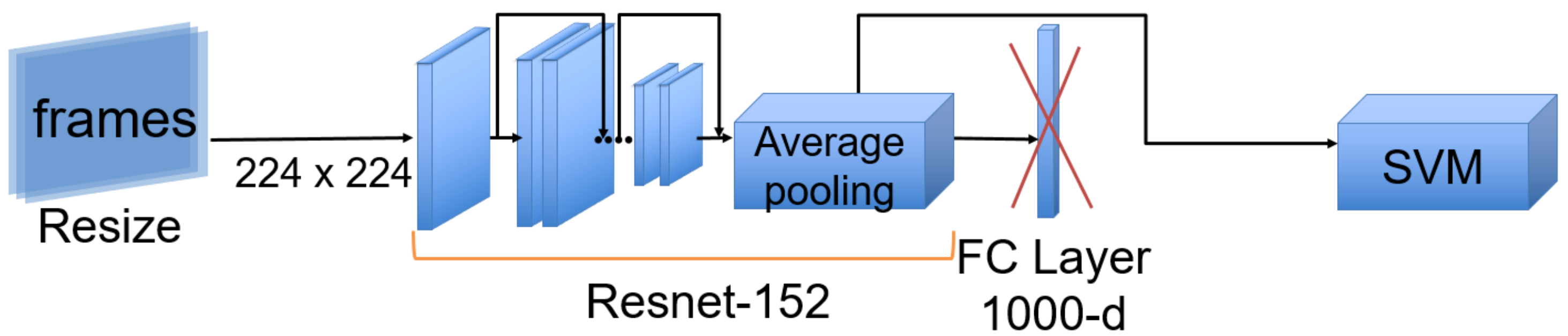}

         \caption{Surface classification model with Resnet-152 + SVM}
         \label{fig:resnet152+svm}
         \vspace{-.5cm}
\end{figure}

\begin{figure}
\vspace{-0.5cm}
\begin{subfigure}[b]{.48\linewidth}
  \centering
  \includegraphics[width=1\linewidth]{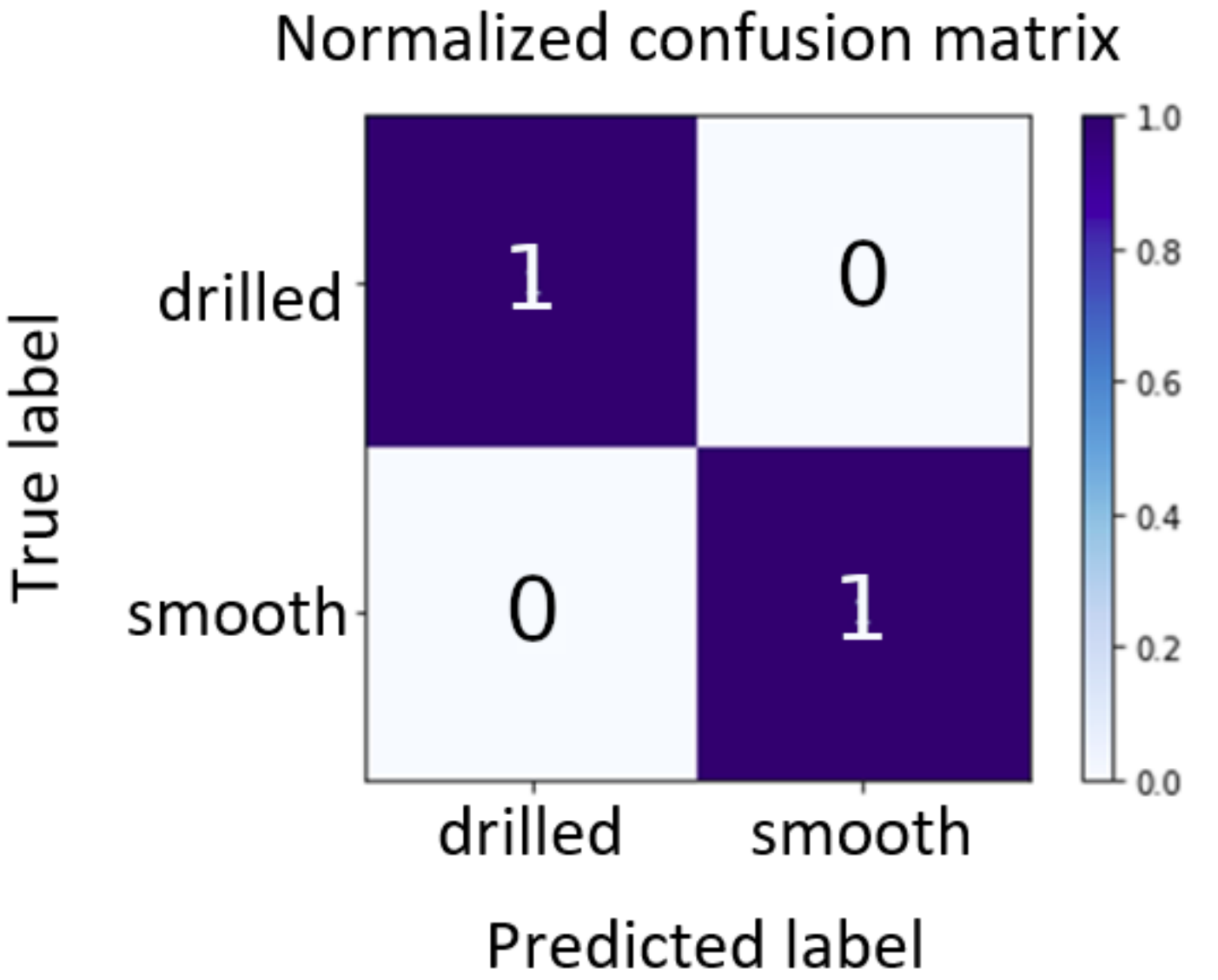}
  \caption{}
  \label{fig:cm_Classification_Model}
\end{subfigure}
\hfill
\begin{subfigure}[b]{.5\linewidth}
  \centering
  \includegraphics[width=.9\linewidth]{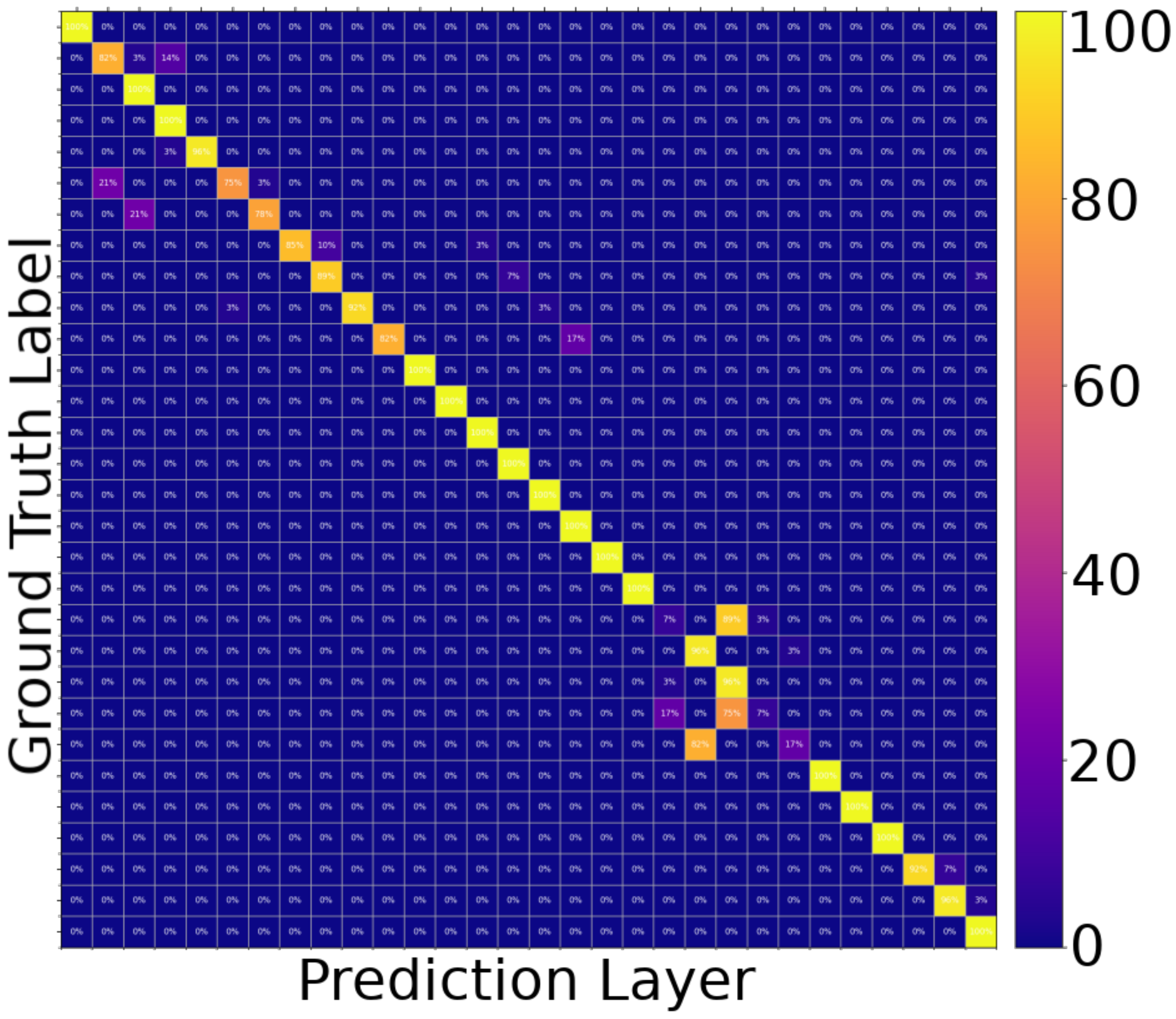}
  \caption{}
  \label{fig:cm_action_recognition_Model}
\end{subfigure}
\vspace{-0.6cm}
\caption{Confusion matrices for the surface classification (a) and action recognition (b) models.}
\end{figure}

\begin{table}[t]
\vspace{-.4cm}
\caption{Comparison analysis on object surface dataset between baseline and proposed model.}
\vspace{-.3cm}
\begin{center}
\begin{tabular}{|l|c|c|}
\hline
\multicolumn{1}{|c|}{\textbf{Model}} & \textbf{Resolution} & \textbf{Accuracy} \\ \hline
Resnet-50                        & 224                   & 90                   \\ \hline
Resnet-152                        & 224                   & 94                   \\ \hline
Resnet-50 + SVM                          & 224                   & $\mathbf{100}$                   \\ \hline
Resnet-152 + SVM                          & 224                   & $\mathbf{100}$                   \\ \hline
\end{tabular}
\end{center}
\label{table:Comparision_resnet_models}
\vspace{-.4cm}
\end{table}

\begin{table}[!h]
\vspace{-.15cm}
\caption{Analysis comparing different approaches on the SlowOnly network for the HRI30 action recognition dataset.}
\vspace{-.1cm}
\centering
\centering
\begin{tabular}{|l|c|c|c|}
\hline
\multicolumn{1}{|c|}{\textbf{Method}} & \textbf{Backbone} & \textbf{Pretrained} & \textbf{Accuracy}  \\ \hline
SlowOnly                              & Resnet-50         & Kinetics-400        & $\mathbf{86.55}$             \\ \hline      
SlowOnly                              & Resnet-50         & Imagenet            & 64.29                                    \\ \hline
\end{tabular}
 \label{table:comparison_model_for_action_recognition}
\vspace{-.5cm}
\end{table}

\begin{figure*}[t]
         \centering
         \includegraphics[ width=1\textwidth]{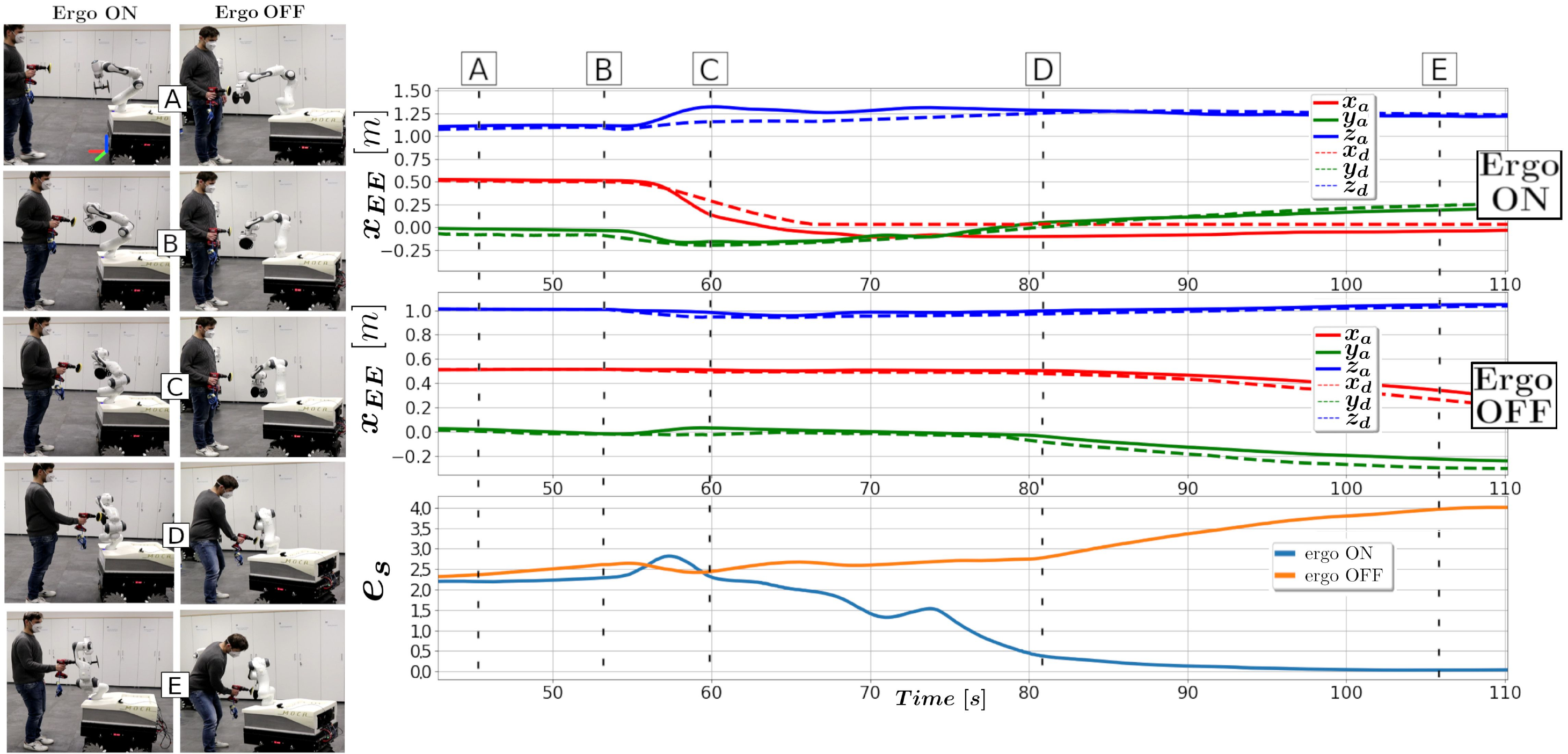}
         \caption{
            Exp. 3. Object reorientation sequences, with (left column) and without (right column) optimal ergonomics \eqref{eq:Hgergo}. The three plots indicate respectively: EE trajectory for both cases with (top) and without (middle) optimal ergonomics, human ergonomics score (bottom).
            }
         \label{fig:exp1}
         \vspace{-0.6cm}
\end{figure*}

\subsection{Action Recognition Model}
\subsubsection{Dataset} \label{subsub:Dataset_action}
The dataset used to train the network is called HRI30 \cite{iodicehri30}. It includes thirty classes associated with actions commonly performed in a collaborative industrial environment. They are of three types: Human-Object Perception, Body-Motion Only, and Human-Robot Collaboration. The action sequences involved 11 subjects performing actions such as delivering the object to the robot, walking, picking up one of the two tools on the table, and interacting with the robot from a stationary position or moving in any direction.

\subsubsection{Description Model} \label{subsub:description_model_action}
We use the SlowOnly neural network for action recognition, i.e., the Slow branch detached from SlowFast \cite{feichtenhofer2019slowfast}, with Resnet50 \cite{he2016deep} as the backbone. It is a network pre-trained on Kinetics-400 \cite{kay2017kinetics}, a dataset that consists of 240k training videos and 20k validation videos across 400 categories of human actions. 

\subsection{3D Human Body Tracking}
To keep track of the human in the workspace, we focus on the 3D reconstruction of the human position.
We extract the 3D position of human body joints by estimating it on the 2D position (u,v) transmitted from the Openpose neural network \cite{cao2017realtime} via RGB-D camera. We simultaneously exploit each pixel's color and depth values of the input RGB-D data, keeping the data organized and filtering out from the point cloud the pixels lacking depth values (NaN) to address occlusion problems arising from the 3D mapping. Finally, we perform the extrapolation by setting the camera to a $640 \times 480$ resolution with a frequency of $20 \si{Hz}$, having a one-to-one correspondence between depth and color data.

%% file: Experiments_revised.tex
To validate the proposed framework, we employ the MObile Collaborative robotic Assistant (MOCA), a collaborative mobile manipulator (m = 6 and n = 10) composed of a Franka Emika Panda robotic arm and a Robotnik SUMMIT-XL STEEL mobile platform. Three Realsense d435i cameras operate at a frequency of $20\si{Hz}$ and a resolution of $640\times 480$.
In Sections \ref{sec:exp1} and \ref{sec:exp2}, we test the performances of the recognition systems. We explain the HRC task in Exp. 3, and in Exp. 4 we allow the human to control the robot by walking ergonomically in space. Finally, in Exp.5, we investigate the subjective workloads reported.

\vspace{-.1cm}

\subsection{\resizebox{0.96\linewidth}{!}{Experiment 1: Surface Classification Model Benchmarking} } \label{sec:exp1}
In this experiment, we classify the surfaces of the objects handed by a human to a robot in the initial phase of the interaction.
We start by splitting the dataset introduced in Sec. \ref{subsub:Dataset} into Train/Val/Test with a ratio of 70/20/10 and, as shown in Fig. \ref{fig:Classification_Model}.C, we preprocess all phases of the input images by using data augmentation techniques, such as scaling $255\times255$ pixels, horizontal flipping, and center cropping crop $224\times224$ pixels, to create virtual copies of different surface images and improve the generalization performance of the model.     
On the dataset, with a learning rate of 0.001 and an SGD (Stochastic Descending Gradient) optimizer, we train Resnet-50 and Resnet-152 to examine neural networks in both the baseline and proposed models.
Firstly (Fig. \ref{fig:resnet152}), we freeze all the weights, i.e., knowledge gained in the pre-training on the ImageNet dataset, and cut the last three layers of the baseline ResNet-50/152, i.e., a fully connected (FC) layer of size 1000, a softmax layer and the output layer. A new FC layer, a new softmax layer, and a new output layer are added. The FC layer of size two is connected to a dropout layer in the transferred ResNet.

In the second case, Fig. \ref{fig:resnet152+svm}, the dataset is used to train the added SVM classifier by applying the same approach introduced in Sec. \ref{subsub:description_model} on the ResNet-152 backbone. Hence, we connect the binary SVM classifier to the trained Average Pooling layer, trained using our dataset of object surface images on the previously described networks via transfer learning.
We report the performance of the different implementations in Table  \ref{table:Comparision_resnet_models}.
The comparison shows that the baseline Resnet-152 outperforms the ResNet-50, given the number of layers that allow for more detailed feature extraction. Instead, the models based on our approach increased in accuracy by $+6\%$ for the ResNet-152 and by $+10\%$ for the Resnet-50 network, reaching $100\%$ accuracy. A representation of this accuracy in class classification is shown in Fig. \ref{fig:cm_Classification_Model}.
Despite the equality of accuracy between the backbones, we choose ResNet-50 as the backbone of our approach for its lower computational and model complexity.

\vspace{-.1cm}

\subsection{Experiment 2: Action Recognition Model Benchmarking} \label{sec:exp2}
Here we analyze the performances of the neural network for action recognition. The dataset used to train the network was introduced in section \ref{subsub:Dataset_action}. It includes 2940 action videos randomly divided into Train and Test with a ratio of 70 and 30.
Before training and testing the network, the input frames are first sampled one every four frames per clip and then resized to 256 $\times$ 256 pixels. Only for the training and tuning phases they are also randomly flipped.
From these settings, we perform comparative performance analysis on different SlowOnly network models by applying a learning rate of 0.001 and a Stochastic Gradient Descent (SGD) optimizer \cite{ruder2016overview} during network training. From the comparison in Table \ref{table:comparison_model_for_action_recognition}, 
we note that the network trained on the HRI30 dataset performs better when pretrained on kinetics-400 instead of ImageNet, and we justify this result by associating it to the different nature of the two pretrained datasets. While the former collects videos of human actions, the latter collects images of objects and animals.
In addition, we use the confusion matrix in Fig. \ref{fig:cm_action_recognition_Model} to analyze the classification performance of each of the classes obtained by training the SlowOnly pre-trained network of Kinetics400 on the HRI30 dataset, from which it can be seen that the model achieved remarkable performance. Here, the diagonal cells represent the percentage of correctly classified test data with their true label, while the other cells represent the percentage of misclassified data. As shown, the model had difficulty
in distinguishing between picking up and putting down the drill, object, or polisher, due to the similarity of actions.
The datasets used in most industrial works are either not publicly available or different from those used in this work, so we cannot compare our results directly.

\begin{figure*}[t!]
         \centering
         \includegraphics[ width=1\textwidth]{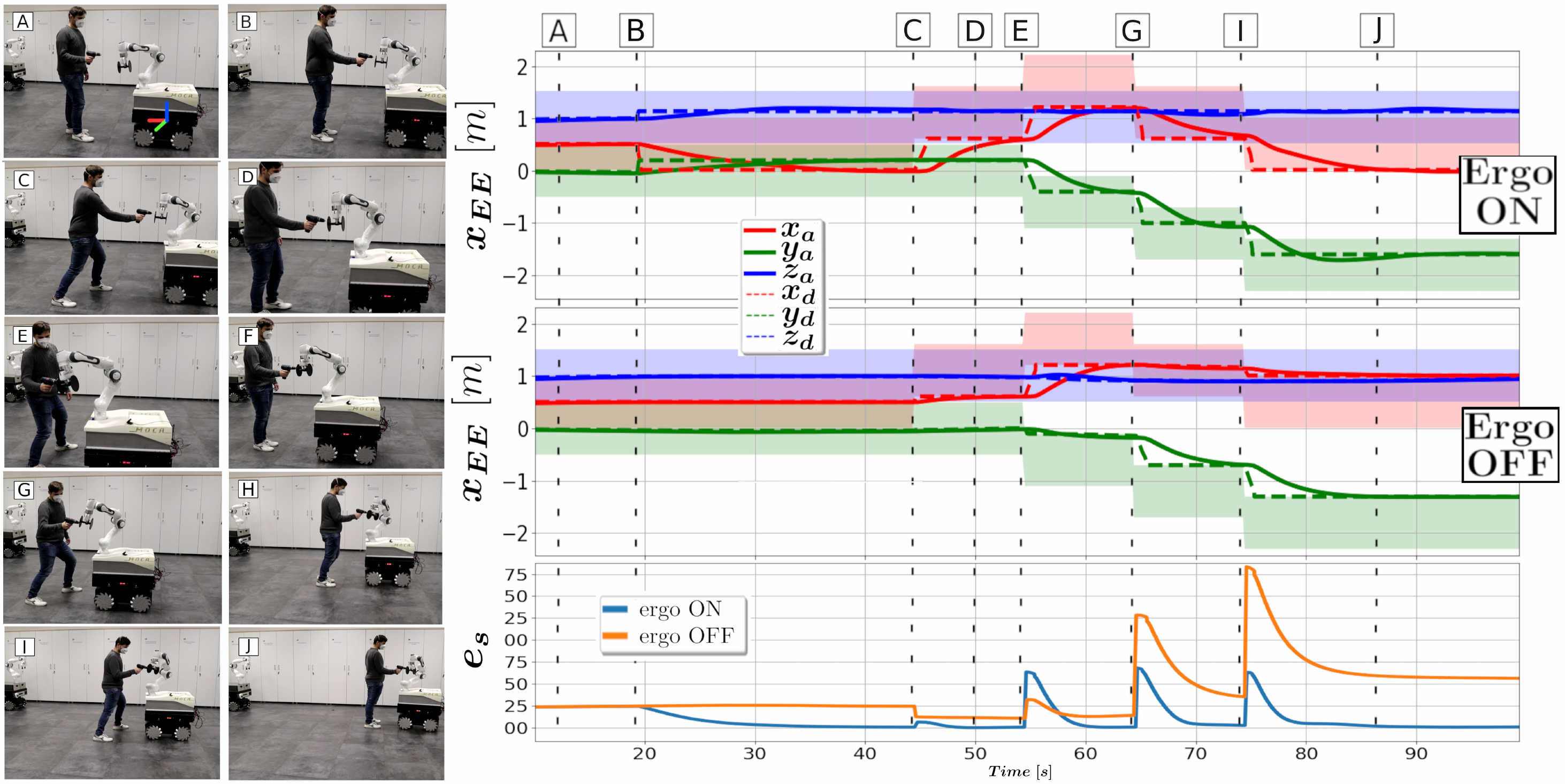}
         \caption{
            Exp. 4. Snapshots sequence dictated by human intentions and actions. Plots: EE trajectory for both cases with (top) and without (middle) optimal ergonomics activation, human ergonomics score (bottom). The coloured regions represent the HRSW along each axis, based on the constraints on $\bm{x}_d$ defined in \eqref{eq:constr_slack}.          
            }
         \label{fig:exp2}
         \vspace{-.5cm}
\end{figure*}

\begin{figure}[ht]
         \centering
         \includegraphics[trim=0cm 0cm 0cm 0cm, width=0.48\textwidth]{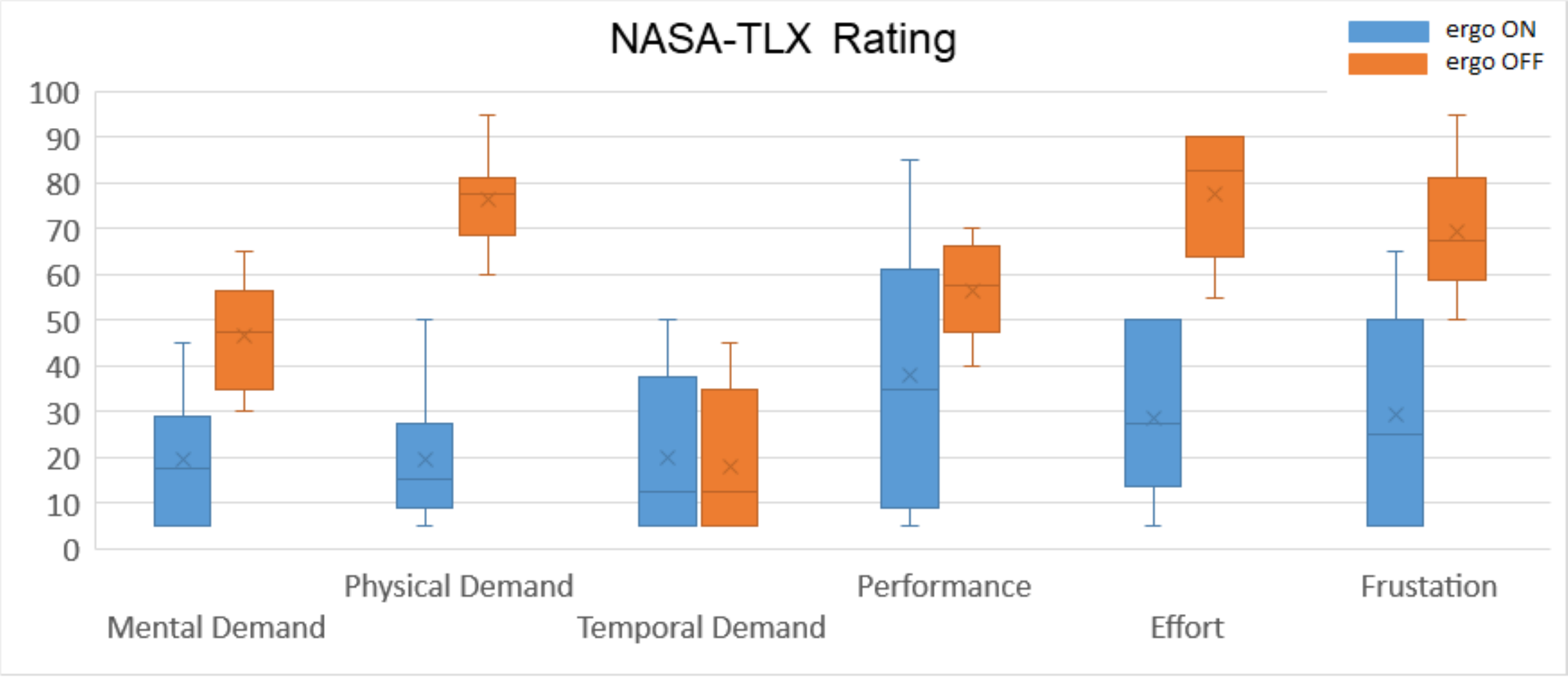}

         \caption{NASA-TLX questionnaire outcomes on HRC task.
            }
         \label{fig:NASA-TLX}
         \vspace{-0.6cm}
\end{figure} 

\vspace{-.1cm}

\subsection{Experiment 3: tasks order dictated by human intentions}
\label{sec:exp3}
In sequence, the subject has to first grab the object of Fig. \ref{fig:Classification_Model} from a table and deliver it to the robot.
The object has two surfaces, each one related to a specific operation to be performed on top (drilling for the perforated surface, polishing for the smooth surface).
Thanks to the action recognition module, as soon as the subject delivers the object by straightening the arm (becoming collaborative), the robot grasps it. It then proceeds with surface identification through the onboard camera, which makes the robot aware of the orientation of the grabbed object, since this will affect the remainder of the experiment. 
Afterwards, based on the tool that the subject decides to use next (either drill or polisher), the robot will adjust the object's orientation, in case the current surface facing the human is not matched with respect to the chosen tool (Fig. \ref{fig:exp1}(A)). Since in the industrial environment, multiple people can carry the same tool, the robot will reorient the object only when the human becomes collaborative, meaning he/she directly engages in the interaction by straightening the arm (Fig. \ref{fig:exp1} (A)). 
Assuming no limiting occlusion, this increases the robustness when other people are in the frame.

If the tool held by the human does not match the current surface, the constraints region of the AHQP, in terms of orientation on $\bm{x_d}$, will be modified through \eqref{eq:constr} in order to rotate the EE as in Fig. \ref{fig:exp1}(C), while optimizing ergonomics. Given the limited range of motion of the robot's wrist and its initial position, a simple rotation of the last joint would be out of reach. Thereby, a full adjustment of the redundancy is necessary, which is correctly achieved by the AHQP. 

To provide a benchmark, we perform the same experiment without the activation of optimal ergonomics (second column snapshots, Fig.\ref{fig:exp1}), thus only complying with the criteria of shortest distance from the actual pose, dictated by
\begin{equation}
\vspace{-.2cm}
    \min_{\bm{\chi}} \| \dot{\bm{x}}_d \|^2,
\end{equation}
which replaces \eqref{eq:Hgergo}.
The bottom plot compares the ergonomics scores of the two cases, showing noticeable improvements, starting from the time instant (B) in which optimal ergonomics is activated. In (D) the human starts interacting, but shortly after (E), the score difference is already large, being $e_s=0.12$ for the proposed controller against $e_s=4.2$, which indicates poor posture and high risk.

\vspace{-.2cm}

\subsection{Experiment 4: Human-Following feature}
\label{sec:exp4}

Proceeding from Exp. 3, while the human is working on the object, he/she might decide to move in space
(Fig. \ref{fig:exp2}), e.g. to reach the next workstation, avoiding idle times. This puts the human in charge of the operation, accommodating movements and intentions during the task. 
Upon becoming collaborative (B), optimal ergonomics is activated, leading to a consistent reduction of $e_s$ with respect to the benchmark.
Thanks to the action recognition module, we can associate the subjects intentions to the movement of their feet. Indeed, during the HRC task, the user can decide to move around the robot by simply starting to walk in space (Fig. \ref{fig:exp2}(C)). Once the action and its direction are detected, the constraints of the AHQP are changed by updating the HRSW towards the new intended direction in \eqref{eq:constr_slack}. 
This is seen in the $\bm{x_{EE}}$ plots of Fig. \ref{fig:exp2}, in which at (C) the first step is detected (human backwards) and the constraints on $\bm{x_{d}}$ \eqref{eq:constr_slack} are updated accordingly (robot forwards). The HRSW (position only) is coloured based on the $x$, $y$, and $z$ contribution, while the actual and desired EE trajectories are depicted with continuous and dashed lines respectively. Thanks to the hierarchical order defined in Sec. \ref{sec:sot}, the EE remains in the feasible region, while optimizing ergonomics. The softened structure allows to exit from the HRSW during transitions, without infeasibility issues.
It is also noticeable how $\bm{x_{d}}$ behaves differently inside of the feasible region for the two cases, e.g., by looking at the $y-$component of $\bm{x_{d}}$, we ensure not only the final solution to lie within the feasible region, but also to optimize ergonomics.
The $e_s$ plot, shows a cumulative increase of the score when ergonomics is inactive, clearly leading to harmful postures.
The online control scheme proposed allows to continuously repeat this process (Fig. \ref{fig:blockscheme}), allowing the robot to smoothly follow the subject.
Hence, while optimally complying with the hierarchy, it is possible to command the robot by detecting the human body's inherent intentions.
\vspace{-.05cm}

\subsection{Experiment 5: Usability evaluation}
\label{sec:exp5}
We here analyze the robustness of our approach across multiple subjects, to assess their perceived workload during Exp.s 3 and 4. This allows to quantify the benefits of the person-specific ergonomic maps, by selecting a proper variety of subjects. Indeed,
ten healthy subjects (7 males and 3 females) conducted the experiments described in Sec.\ref{sec:exp3} and Sec.\ref{sec:exp4}. Their average age is $28.5$ years, and their heights range from $ 1.58 \si{m}$ to $ 1.94\si{m}$. Without person-specific maps, such gap ($0.36\si{m}$) would clearly negatively impact the subjects experience, affecting the final scores. An initial training phase allowed the subjects to gain familiarity with the experiment, during which they expressed appreciation for the lack of uncomfortable wearable devices, that required initial calibration. Following the experiment, each subject was asked to fill out a NASA-TLX questionnaire form to assess how the experience was perceived. This is composed of the following scales scored from 0 to 20: temporal demand  (TD), mental demand (MD), effort (EF), physical demand (PD), performance (PE), and frustration (FR).
Fig. \ref{fig:NASA-TLX} shows improved overall PE (inverted scale), highlighting the ease of use of our framework in dynamic HRC. MD and PD are reduced as well, highlighting better execution responsiveness, and lower effort and stress levels.
The data collection was carried out in accordance with the Declaration of Helsinki, and the protocol was approved by the ethics committee Azienda Sanitaria Locale (ASL) Genovese N.3 (Protocol IIT\_HRII\_ERGOLEAN 156/2020).

\vspace{-.1cm}

%% file: Conclusion_revised.tex

The proposed framework identifies human actions and intentions through both surface classification and action recognition models. By formulating a hierarchical controller based on AHQP, we close the loop on the human figure, creating a framework that allows the operator to control the robot's redundancy by simply moving in space and letting the robot follow accordingly. The gathered data indicates noticeable ergonomics improvements, essential to reduce MSDs for future applications and further studies.

Possible limitations, besides common camera occlusion problems, involve the typical limitations of HQP when dynamically altering the stack of tasks. Indeed, optimal ergonomics must be carefully shifted in the hierarchy and the issue of continuity must be addressed in future studies.
\vspace{-.2cm}

%% file: main_revised.bbl
\begin{thebibliography}{10}
\providecommand{\url}[1]{#1}
\csname url@samestyle\endcsname
\providecommand{\newblock}{\relax}
\providecommand{\bibinfo}[2]{#2}
\providecommand{\BIBentrySTDinterwordspacing}{\spaceskip=0pt\relax}
\providecommand{\BIBentryALTinterwordstretchfactor}{4}
\providecommand{\BIBentryALTinterwordspacing}{\spaceskip=\fontdimen2\font plus
\BIBentryALTinterwordstretchfactor\fontdimen3\font minus
  \fontdimen4\font\relax}
\providecommand{\BIBforeignlanguage}[2]{{%
\expandafter\ifx\csname l@#1\endcsname\relax
\typeout{** WARNING: IEEEtran.bst: No hyphenation pattern has been}%
\typeout{** loaded for the language `#1'. Using the pattern for}%
\typeout{** the default language instead.}%
\else
\language=\csname l@#1\endcsname
\fi
#2}}
\providecommand{\BIBdecl}{\relax}
\BIBdecl

\bibitem{ajoudani2018progress}
A.~Ajoudani, A.~M. Zanchettin, S.~Ivaldi, A.~Albu-Sch{\"a}ffer, K.~Kosuge, and
  O.~Khatib, ``Progress and prospects of the human--robot collaboration,''
  \emph{Autonomous Robots}, vol.~42, no.~5, pp. 957--975, 2018.

\bibitem{xiong2020transferable}
Q.~Xiong, J.~Zhang, P.~Wang, D.~Liu, and R.~X. Gao, ``Transferable two-stream
  convolutional neural network for human action recognition,'' \emph{Journal of
  Manufacturing Systems}, vol.~56, pp. 605--614, 2020.

\bibitem{liu2019deep}
Z.~Liu, Q.~Liu, W.~Xu, Z.~Liu, Z.~Zhou, and J.~Chen, ``Deep learning-based
  human motion prediction considering context awareness for human-robot
  collaboration in manufacturing,'' \emph{Procedia CIRP}, vol.~83, pp.
  272--278, 2019.

\bibitem{feichtenhofer2019slowfast}
C.~Feichtenhofer, H.~Fan, J.~Malik, and K.~He, ``Slowfast networks for video
  recognition,'' in \emph{Proceedings of the IEEE/CVF international conference
  on computer vision}, 2019, pp. 6202--6211.

\bibitem{tran2018closer}
D.~Tran, H.~Wang, L.~Torresani, J.~Ray, Y.~LeCun, and M.~Paluri, ``A closer
  look at spatiotemporal convolutions for action recognition,'' in
  \emph{Proceedings of the IEEE conference on Computer Vision and Pattern
  Recognition}, 2018, pp. 6450--6459.

\bibitem{zhang2020recurrent}
J.~Zhang, H.~Liu, Q.~Chang, L.~Wang, and R.~X. Gao, ``Recurrent neural network
  for motion trajectory prediction in human-robot collaborative assembly,''
  \emph{CIRP annals}, vol.~69, no.~1, pp. 9--12, 2020.

\bibitem{Mansard2009_sot}
N.~Mansard, O.~Khatib, and A.~Kheddar, ``A unified approach to integrate
  unilateral constraints in the stack of tasks,'' \emph{IEEE Transactions on
  Robotics}, vol.~25, no.~3, pp. 670--685, 2009.

\bibitem{Khatib2008}
O.~Khatib, L.~Sentis, and J.-H. Park, ``A unified framework for whole-body
  humanoid robot control with multiple constraints and contacts,'' in
  \emph{European Robotics Symposium 2008}.\hskip 1em plus 0.5em minus
  0.4em\relax Springer, 2008, pp. 303--312.

\bibitem{Tassi_Gholami2021}
S.~Gholami, F.~Tassi, E.~De~Momi, and A.~Ajoudani, ``A reconfigurable interface
  for ergonomic and dynamic tele-locomanipulation,'' in \emph{2021 IEEE/RSJ
  International Conference on Intelligent Robots and Systems (IROS)}, 2021, pp.
  4260--4267.

\bibitem{Chenguang_Demomi}
H.~Su, C.~Yang, G.~Ferrigno, and E.~De~Momi, ``Improved human–robot
  collaborative control of redundant robot for teleoperated minimally invasive
  surgery,'' \emph{IEEE Robotics and Automation Letters}, vol.~4, no.~2, pp.
  1447--1453, 2019.

\bibitem{Escande}
A.~Escande, N.~Mansard, and P.-B. Wieber, ``Hierarchical quadratic programming:
  Fast online humanoid-robot motion generation,'' \emph{The International
  Journal of Robotics Research}, vol.~33, no.~7, pp. 1006--1028, 2014.

\bibitem{Kanoun}
O.~Kanoun, F.~Lamiraux, and P.-B. Wieber, ``Kinematic control of redundant
  manipulators: Generalizing the task-priority framework to inequality task,''
  \emph{IEEE Transactions on Robotics}, vol.~27, no.~4, pp. 785--792, 2011.

\bibitem{Tassi2021}
F.~Tassi, E.~De~\hspace{.1mm} Momi, and A.~Ajoudani, ``Augmented hierarchical
  quadratic programming for adaptive compliance robot control,'' in \emph{2021
  IEEE International Conference on Robotics and Automation (ICRA)}, 2021, pp.
  3568--3574.

\bibitem{TassiRCIM}
\BIBentryALTinterwordspacing
F.~Tassi, E.~De~Momi, and A.~Ajoudani, ``An adaptive compliance hierarchical
  quadratic programming controller for ergonomic human-robot collaboration,''
  \emph{Robotics and Computer-Integrated Manufacturing}, 2022. [Online].
  Available: \url{tinyurl.com/464d9tzs}
\BIBentrySTDinterwordspacing

\bibitem{Busch2018}
B.~Busch, M.~Toussaint, and M.~Lopes, ``Planning ergonomic sequences of actions
  in human-robot interaction,'' in \emph{IEEE International Conference on
  Robotics and Automation (ICRA)}, 2018.

\bibitem{Heydaryan_HRC_ergo}
S.~Heydaryan, J.~Suaza~Bedolla, and G.~Belingardi, ``Safety design and
  development of a human-robot collaboration assembly process in the automotive
  industry,'' \emph{Applied Sciences}, vol.~8, no.~3, 2018.

\bibitem{Wansoo}
W.~Kim, M.~Lorenzini, P.~Balatti, Y.~Wu, and A.~Ajoudani, ``Towards ergonomic
  control of collaborative effort in multi-human mobile-robot teams,'' in
  \emph{2019 IEEE/RSJ International Conference on Intelligent Robots and
  Systems (IROS)}, 2019, pp. 3005--3011.

\bibitem{he2016deep}
K.~He, X.~Zhang, S.~Ren, and J.~Sun, ``Deep residual learning for image
  recognition,'' in \emph{Proceedings of the IEEE conference on computer vision
  and pattern recognition}, 2016, pp. 770--778.

\bibitem{iodicehri30}
\BIBentryALTinterwordspacing
F.~Iodice, E.~De~Momi, and A.~Ajoudani, ``Hri30: An action recognition dataset
  for industrial human-robot interaction,'' 2022. [Online]. Available:
  \url{tinyurl.com/2p9y5jrz}
\BIBentrySTDinterwordspacing

\bibitem{kay2017kinetics}
W.~Kay, J.~Carreira, K.~Simonyan, B.~Zhang, C.~Hillier, S.~Vijayanarasimhan,
  F.~Viola, T.~Green, T.~Back, P.~Natsev \emph{et~al.}, ``The kinetics human
  action video dataset,'' \emph{arXiv preprint arXiv:1705.06950}, 2017.

\bibitem{cao2017realtime}
Z.~Cao, T.~Simon, S.-E. Wei, and Y.~Sheikh, ``Realtime multi-person 2d pose
  estimation using part affinity fields,'' in \emph{Proceedings of the IEEE
  conference on computer vision and pattern recognition}, 2017.

\bibitem{ruder2016overview}
S.~Ruder, ``An overview of gradient descent optimization algorithms,''
  \emph{arXiv preprint arXiv:1609.04747}, 2016.

\end{thebibliography}
